\documentclass[runningheads]{llncs}
\usepackage[T1]{fontenc}
\usepackage{graphicx,verbatim}
\usepackage{graphicx,verbatim}
\usepackage{amsmath}
\usepackage{color}
\usepackage{xcolor,colortbl}
\usepackage{epstopdf}
\usepackage{tabularx,threeparttable}
\usepackage{booktabs}
\usepackage{amssymb}
\usepackage{multirow}
\usepackage[hyphens]{url}  
\usepackage{breakurl}      
\begin{document}
\title{MAKE: Multi-Aspect Knowledge-Enhanced Vision-Language Pretraining for Zero-shot Dermatological Assessment}
%

\author{
Siyuan Yan$^{1,2}$\thanks{Equal contribution}\quad
Xieji Li$^{2*}$\quad
Ming Hu$^{1,2}$\quad
Yiwen Jiang$^{2}$ \quad
Zhen Yu$^{2}$\quad\\
Zongyuan Ge$^{2}$\quad\\
}

\institute{$^1$Faculty of Engineering, Monash University, Melbourne, Australia \\
$^2$ AIM for Health Lab, Monash University, Victoria, Australia}

\maketitle              
\begin{abstract} Dermatological diagnosis represents a complex multimodal challenge that requires integrating visual features with specialized clinical knowledge. While vision-language pretraining (VLP) has advanced medical AI, its effectiveness in dermatology is limited by text length constraints and the lack of structured texts. In this paper, we introduce MAKE, a \textbf{M}ulti-\textbf{A}spect \textbf{K}nowledge-\textbf{E}nhanced vision-language pretraining framework for zero-shot dermatological tasks. Recognizing that comprehensive dermatological descriptions require multiple knowledge aspects that exceed standard text constraints, our framework introduces: (1) a multi-aspect contrastive learning strategy that decomposes clinical narratives into knowledge-enhanced sub-texts through large language models, (2) a fine-grained alignment mechanism that connects sub-captions with diagnostically relevant image features, and (3) a diagnosis-guided weighting scheme that adaptively prioritizes different sub-captions based on clinical significance prior. Through pretraining on 403,563 dermatological image-text pairs collected from education resources, MAKE significantly outperforms state-of-the-art VLP models on eight datasets across zero-shot skin disease classification, concept annotation, and cross-modal retrieval tasks. Our code will be made publicly available at \url{https://github.com/SiyuanYan1/MAKE}. 

\keywords{Dermatology  \and Vision-language \and Knowledge augmentation.}
\end{abstract}
\section{Introduction} 

Dermatological diagnosis represents a complex multimodal challenge \cite{liu2020deep}, requiring clinicians to simultaneously interpret visual features of skin lesions alongside patient history and various clinical concept descriptions \cite{daneshjou2022skincon,monet,trust_derm} for accurate assessment. While deep learning has advanced automated diagnosis systems \cite{derm}, standard approaches face significant limitations when applied to dermatology. Specifically, supervised \cite{derm,liu2020deep} and self-supervised techniques \cite{panderm,swavderm} require extensive labeled data for different tasks and struggle to capture the rich, multimodal nature of dermatological knowledge. This multimodal complexity necessitates more sophisticated approaches that can effectively bridge visual and linguistic understanding for dermatological diagnosis. 
\begin{figure*}[t]
\begin{center}
\includegraphics[width=0.8\linewidth]{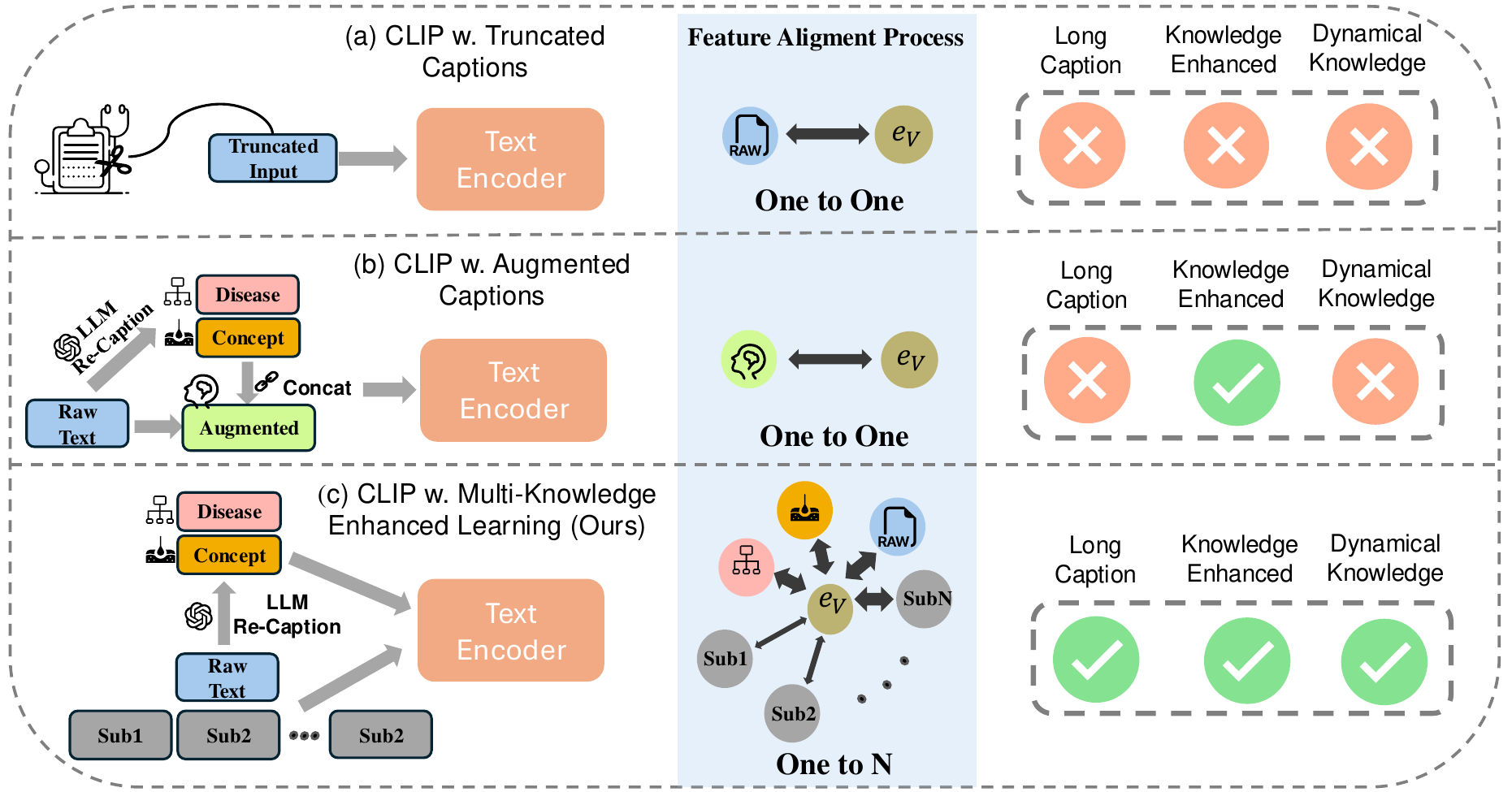}
\vspace{-2mm}
\caption{\textbf{Comparison between training strategies. }Our framework utilizes CLIP with multi-knowledge enhanced learning, addressing long caption modeling limitations. It enables dynamic knowledge modeling between each image and its multiple corresponding captions, each capturing diverse aspects of crucial dermatological  knowledge.}
\label{fig:motivation}
\end{center}
\end{figure*}

In parallel, vision-language pretraining (VLP) \cite{clip} has emerged as a powerful paradigm for multimodal tasks, demonstrating remarkable generalization capabilities without extensive fine-tuning, a capability formally known as zero-shot learning. By learning from rich textual information rather than single-label annotations, these models offer a way to alleviate the heavy dependence on labeled data. However, applying VLP models in dermatology faces two significant challenges. First, conventional VLP frameworks like CLIP \cite{clip} typically limit text input to a fixed token length (e.g., 77 tokens), truncating any longer description (Fig~\ref{fig:motivation}a). This truncation oversimplifies the rich clinical narrative and discards vital diagnostic details, ultimately constraining the model's ability to capture the nuanced clinical concepts of skin lesions. Second, dermatology lacks standardized image-text pairs that are crucial for effective VLP training. Unlike other medical specialties such as radiology \cite{mimic-cxr,chexpert} where structured reports provide well-organized image-text pairs, dermatology often relies on unstructured clinical narratives. Recent works \cite{quilt1m,monet,ophclip} attempt to crawl image-text pairs from web sources, which frequently yield noisy text. Some approaches \cite{laclip,augpair} have explored knowledge augmentation via large language models (LLMs) to generate more comprehensive knowledge-enhanced captions, as illustrated in Fig.~\ref{fig:motivation}b. Yet, VLP models trained using these methods still struggle to model the complex interrelationships among multiple aspects of clinical knowledge—such as lesion morphology, standardized disease descriptions, and associated symptoms—within a single short-length description. Additionally, existing methods \cite{dreamlip} treat all information equally, neglecting the varying contributions of different aspects of knowledge to the final diagnosis. 

To overcome these dermatology-specific challenges, we propose MAKE, a \textbf{M}ulti-\textbf{A}spect \textbf{K}nowledge-\textbf{E}nhanced vision-language pretraining framework sp-ecifically designed for zero-shot dermatological tasks (Fig.~\ref{fig:motivation}c). Our framework introduces three key innovations: First, a multi-aspect knowledge-image contrastive learning strategy that decomposes complex dermatological descriptions into multiple sub-captions, each capturing distinct aspects of clinical knowledge such as morphology, distribution patterns, and associated symptoms. This approach not only mitigates the text length constraint but also enables precise alignment between visual features of skin lesions and various aspects of clinical knowledge, critical for differential diagnosis. Second, a fine-grained alignment mechanism that associates multiple sub-captions with diagnostically relevant image patches of skin lesions, enabling different aspects of dermatological knowledge to jointly characterize the salient visual features crucial for accurate diagnosis. Third, a diagnosis-guided weighting scheme that adaptively prioritizes different aspects of knowledge based on their diagnostic relevance in dermatology practice, better reflecting how dermatologists assign varying importance to different clinical attributes during the diagnostic reasoning process. 

In summary, our contributions include: (1) introducing MAKE, the first vision-language pretraining framework for dermatology; (2) proposing three complementary technical innovations described above that enable fine-grained knowle-dge-enhanced visual-textual learning for dermatological applications; and (3) verifying our method through pretraining on 403,563 dermatological image-text pairs. Through extensive experiments, we demonstrate that MAKE significantly outperforms state-of-the-art VLP models on zero-shot skin disease classification, concept annotation, and cross-modal retrieval tasks across eight datasets.
\begin{figure*}[t]
\begin{center}
\includegraphics[width=0.9\linewidth]{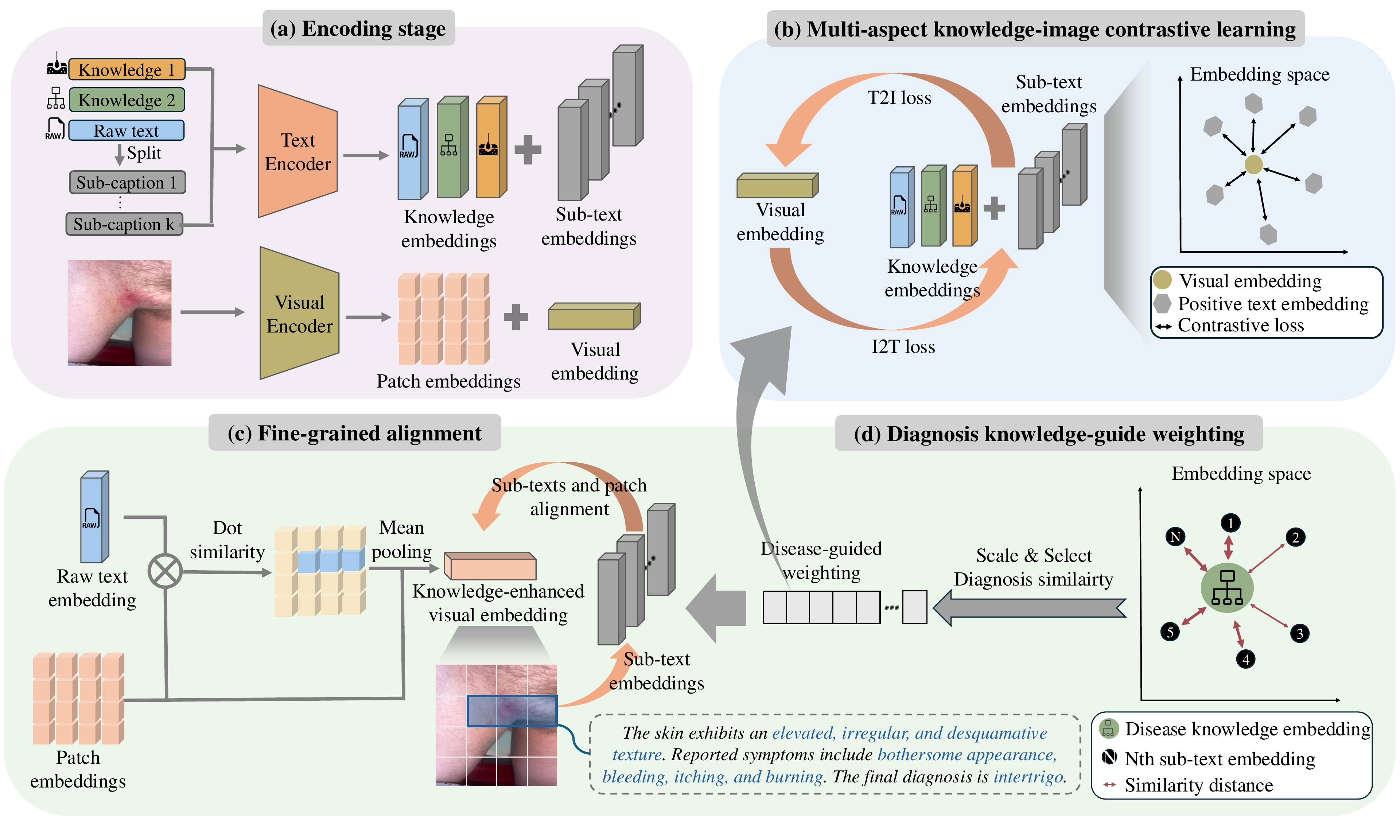}
\vspace{-3mm}
\caption{\textbf{Overview of our MAKE framework.} (a) Encoding multi-aspect clinical knowledge. (b) The process of aligning visual embeddings with multiple positive text embeddings. \textbf{Knowledge 1}: \textit{disease-aspect text}; \textbf{Knowledge 2}: \textit{concept-aspect text}. (c) A fine-grained alignment process, which matches each subtext embedding with knowledge-enhanced visual embeddings. (d) Diagnosis similarity-based weights that modulate alignment between subtexts and visual embeddings.}
\label{fig:make_framework}
\end{center}
\end{figure*}
\section{Methodology}
As illustrated in Fig.~\ref{fig:make_framework}, our MAKE framework comprises three core components: multi-aspect knowledge-image contrastive learning, fine-grained alignment, and diagnosis knowledge-guided weighting. We detail each of them below.
\subsection{Encoding Stage} \label{2.1}
The original dataset consists of image-text pairs $\mathcal{D} = \{(I_i, T_i^{r})\}_{i=1}^M$, where $I_i$ denotes the $i$-th skin image and $T_i^{r}$ represents its associated raw text description. Traditional VLMs like CLIP \cite{clip} are constrained by text length limitations and cannot effectively leverage the rich clinical knowledge in these descriptions.

To address this limitation, we expand each image-text pair $(I_i, T_i^{r})$ into a richer multi-aspect representation through two complementary augmentation methods: 
1) \textbf{Knowledge extraction}: We use LLMs to extract and generate two specialized knowledge aspects from the original text:
   - \textit{Disease aspect text} ($t_i^d$): Contains standardized disease terminology, synonyms, and hierarchical relationships (e.g., melanoma's superclass is ``malignant'').
   - \textit{Concept aspect text} ($t_i^c$): Captures interpretable clinical descriptors from dermatological lexicons (e.g., ``plaque'', ``scale'', ``erosion'') crucial for diagnosis. 
2) \textbf{Sentence decomposition}: We preserve the raw text's detailed content by splitting $T_i^{r}$ into multiple sentences, creating a subtext set $S_i = \{S_i^j\}_{j=1}^K$ of size $K$ focusing on different aspects of the raw description. To this end, we obtain two complementary sets: a \textit{knowledge set} $\{T_i^{r}, t_i^d, t_i^c\}$ containing the raw text and derived knowledge aspects, and a \textit{subtext set} $S_i$ from sentence decomposition.

Given a batch $\mathcal{B} = \{I_i, T_i^{r}\}_{i=1}^N$ of $N$ samples, our framework transforms each image-text pair into an enhanced representation:
\begin{equation}
    \mathcal{B}_{\text{enhanced}} = \{I_i, (T_i^r, t_i^d, t_i^c, \{S_i^j\}_{j=1}^K)\}_{i=1}^N
\end{equation}
This approach overcomes text length constraints while capturing multi-aspect clinical knowledge.

Then, we process inputs through dedicated encoders. Using the vision encoder $E_V$, we obtain the normalized visual embedding $e_i^v = E_V(I_i)$ and patch embeddings $e_i^p = [v_i^1, ..., v_i^{HW}]$. For textual part, we employ the text encoder $E_T$ to project each subtext from both the \textit{knowledge set} and \textit{subtext set} individually, resulting in $K+3$ text embeddings:
\begin{equation}
\label{eq2}
e_i^t =(e^r_i, e^d_i, e^c_i, \{e^{s_j}_i\}_{j=1}^K) =(E_T(T_i^r), E_T(t_i^d), E_T(t_i^c), \{E_T(S_i^j)\}_{j=1}^K)  
\end{equation}
where $K$ is the number of subtexts from \textit{subtext set}, and 3 represents the embeddings of raw, \textit{disease aspect}, and \textit{concept aspect text} from \textit{knowledge set}.

\subsection{Multi-aspect Knowledge-Image Contrastive Learning} \label{2.2}
To optimize multiple aspect texts with their corresponding single image, we apply multi-positive contrastive learning \cite{laclip}, as shown in Fig.~\ref{fig:make_framework}(b). We align the visual embedding with all associated $K+3$ text embeddings from both the \textit{knowledge set} and \textit{subtext set} in the shared embedding space using an image-to-text multi-positive contrastive learning loss:
\begin{equation}
\label{equ: mkcl_i2t}
\mathcal{L}_{i2t}^{mkcl} = -\sum_{i=1}^N\sum_{j=1}^{K+3}{log\frac{exp(sim\langle e_i^v, e_{ij}^t \rangle / \tau)}{\sum_{n=1}^N{exp(sim\langle e^v_n, e^{t}_{ij} \rangle/\tau)}}}
\end{equation}
where $e_{ij}^t$ represents the $j$-th text embedding of the $i$-th sample with $j \in \{r,d,c,s_1,...,s_K\}$ as defined in Eq.~\ref{eq2}, $sim\langle .,. \rangle$ denotes cosine similarity, and $\tau$ is a learnable temperature parameter. Similarly, the text-to-image loss is:

\begin{equation}
\label{equ: mkcl_t2i}
\mathcal{L}_{t2i}^{mkcl} = -\sum_{i=1}^N\sum_{j=1}^{K+3}{log\frac{exp(sim\langle e_{ij}^t, e_i^v \rangle / \tau)}{\sum_{n=1}^N{exp(sim\langle e_{ij}^t, e_n^v \rangle/\tau)}}}
\end{equation}

The final multi-aspect knowledge-image contrastive learning loss is defined as $\mathcal{L}^{mkcl} = (\mathcal{L}_{t2i}^{mkcl}+\mathcal{L}_{i2t}^{mkcl})/2$.

\subsection{Fine-grained Alignment} \label{2.3}
To enhance the fine-grained alignment capability of VLMs, we draw inspiration from dermatologists who leverage multiple knowledge aspects to characterize a skin lesion. As shown in Fig.~\ref{fig:make_framework}(c), we align all subtexts from \textit{subtext set} with specific knowledge-enhanced patches to improve fine-grained alignment.

Specifically, we first calculate dot product similarity between patch embeddings $e_i^p = [v_i^1, \dots, v_i^{HW}]$ and raw text embedding $e_i^r$ to generate a normalized similarity map $ z_i = e_i^r \cdot (e_i^p)^T$. Next, we compute the dot product between this similarity map and patch embeddings to highlight patches with strong knowledge-semantic relevance. Finally, we apply mean pooling to align the dimensionality of knowledge-enhanced visual embeddings with sub-caption embeddings. The knowledge-enhanced visual embedding is formulated as:
\begin{equation} 
\label{equ: knowledge_enhanced_embedding} 
e^k_i = \sum_{n=1}^{HW} v_i^n \cdot \frac{z_i^n}{\sum_{j=1}^{HW}z_i^j}  
\end{equation}
where $z_i^n$ is the $n$-th element of the similarity map $z_i$, $v_i^n$ represents the $n$-th patch embedding, and $HW$ is the total number of image patches.

To align each subtext embedding with the knowledge-enhanced visual embedding, we define the fine-grained alignment loss as:
\begin{equation}
\label{equ: slra}
\mathcal{L}_{slra} = -\sum_{i=1}^N\sum_{j=1}^{K}{log\frac{exp(sim\langle e^{s_j}_i, e_i^k \rangle / \tau)}{\sum_{n=1}^N{exp(sim\langle e^{s_j}_i, e_n^k \rangle/\tau)}}}
\end{equation}
where $e^{s_j}_i$ denotes the embedding of the $j$-th subtext for the $i$-th sample and $e_i^k$ is the knowledge-enhanced visual embedding from Eq.~\ref{equ: knowledge_enhanced_embedding}.

\subsection{Diagnosis Knowledge-guided Weighting} \label{2.4}
Mimicking how dermatologists prioritize clinical information, our approach adaptively weights text elements by diagnostic relevance. As shown in Fig.~\ref{fig:make_framework}(d), we introduce a weighting mechanism reflecting clinical decision-making. For each sample, we compute subtext weights for \textit{subtext set} by measuring semantic similarity between subtext embeddings $\{e^{s_j}_i\}_{j=1}^{K}$ and \textit{disease aspect} embedding $e^d_i$:
\begin{equation}
\label{equ:disease_guided_weight}
w_i = \frac{\{e^d_i \cdot (e^{s_j}_i)^T\}_{j=1}^{K}}{\max(\{e^d_i \cdot (e^{s_j}_i)^T\}_{j=1}^{K})}
\end{equation}
This yields weight vector $w_i = [w_i^1, ..., w_i^K]$ for each sample's $K$ subtexts, normalized by maximum similarity. Batch-wide weights are denoted as $\hat{w} = \{w_i\}_{i=1}^N$. \textit{Knowledge set} embeddings (raw text, disease, and concept aspects) receive default weights of 1 as they already contain rich diagnostic information.
Our final loss integrates contrastive and fine-grained alignment losses, modulated by these diagnosis-guided weights.
\begin{equation}
\label{equ:total_loss}
\mathcal{L}_{total} = \hat{w}_{mkcl}\mathcal{L}_{mkcl} + \lambda \hat{w}_{slra}\mathcal{L}_{slra}
\end{equation}
where $\lambda$ balances the two loss terms, while $\hat{w}_{mkcl}$ and $\hat{w}_{slra}$ represent the weights applied to each respective loss component derived from Eq.~\ref{equ:disease_guided_weight}.
\setlength
\tabcolsep{2.5pt}
\section{Experimental details}

\noindent\textbf{Experiment Setup:} We conduct pretraining on Derm1M \cite{derm1m}, a dataset of 403,563 skin image-text pairs, including 100,487 pairs from PubMed and medical textbooks following the data crawling process of \cite{monet}, with remaining data from YouTube and Twitter sources as in \cite{quilt1m}. We denote Derm1M$^\dagger$ as our knowledge-augmented version where \textit{disease aspect} and \textit{concept aspect texts} are pre-pended to raw text. For evaluation, we use eight downstream datasets (PAD \cite{PACHECO2020106221}, DermNet \cite{Dermnet}, Fitzpatrick17K \cite{groh2021evaluating}, SD-128 \cite{10.1007/978-3-319-46466-4_13}, SNU-134 \cite{Han2019}, SkinCon \cite{daneshjou2022skincon}, Derm7pt \cite{Kawahara2018-7pt}, and SkinCAP \cite{skincap}) across three categories: (1) Zero-shot disease classification for skin cancer and general skin condition diagnosis; (2) Zero-shot concept annotation for identifying clinically relevant concepts that aid diagnosis and interpretability \cite{monet,daneshjou2022skincon}; and (3) Cross-modal retrieval for both image-to-text and text-to-image retrieval. Following CLIP \cite{clip}, we employ zero-shot evaluation without fine-tuning.\\

\noindent\textbf{Implementation Details:} Following CLIP \cite{clip}, we use ViT-B/16 \cite{vit} as the image encoder and GPT2 \cite{gpt2} with a context length of 77 as the text encoder. Our proposed MAKE leverages all three text types (raw, \textit{disease aspect}, and \textit{concept aspect text}) of Derm1M$^\dagger$. To ensure a fair comparison with state-of-the-art VLM methods, we trained each baseline model in two configurations: one using only raw text (denoted as training on the Derm1M dataset) and another using knowledge-augmented text (denoted as training on the Derm1M$^\dagger$ dataset). Models are trained for 15 epochs with a batch size of 2048, a learning rate of $1e-4$, and a 1500-step warm-up with a weight decay of 0.1. The loss weighting factor $\lambda$ is 0.7, and images are processed at $224 \times 224$ resolution. For all models, we use the final checkpoint and conduct extensive hyperparameter tuning to find the optimal model. 
\section{Results}
\begin{table}[t]
    \centering
     \small 
    \caption{Zero-shot performance comparison for disease and concept classification.}
      \vspace{-2mm}
    \resizebox{1\textwidth}{!}{
    \begin{tabular}{l|c|cccccc|ccc}
        \toprule
        \multirow{2}{*}{\textbf{Method}} & \multirow{2}{*}{\textbf{Pretrain Data}} & \multicolumn{6}{c|}{\textbf{Disease Classification(ACC)}} & \multicolumn{3}{c}{\textbf{Concept Annotation(AUROC)}}\\
        &  & \textbf{PAD} & \textbf{DermNet} & \textbf{F17K} & \textbf{SD-128} & \textbf{SNU-134} & \textbf{Average} & \textbf{SkinCon} & \textbf{Derm7pt} & \textbf{Average}\\
        \hline
        \textbf{Class number} &   & 6 & 23 & 114 & 128 & 134 &  & 32 & 7 & \\
        \textbf{Test size} &  & 2,298 & 19,559 & 16,577 & 5,619 & 2,101 &  & 3,855 & 1,011 & \\ 
        \hline
        CLIP \cite{clip} & CLIP400M & 0.4330 & 0.1738 & 0.0628 & 0.0728 & 0.0733 & 0.1631 & 0.6642 & 0.5594 & 0.6118\\
        BiomedCLIP \cite{biomedclip} & PMC-15M & 0.4295 & 0.1954 & 0.0890 & 0.1321 & 0.0971 & 0.1886 & 0.6817 & 0.6092 & 0.6455 \\
        PMC-CLIP \cite{pmc-clip} & PMC-OA & 0.4312 & 0.1458 & 0.0268 & 0.0443 & 0.0390 & 0.1374 & 0.6251 & 0.5820 & 0.6036 \\
        MONET \cite{monet} & PubMed+TextBook & 0.4308 & 0.2304 & 0.1409 & 0.2072 & 0.1333 & 0.2285 & 0.7502 & 0.6889 & 0.7196\\
        \hline
      
        CLIP \cite{clip} &  Derm1M  & \textbf{0.5957} & 0.7508 & 0.2595 & 0.3349 & 0.2689 & 0.4420 & 0.7233 & 0.6707 & 0.6970 \\
                SigLIP \cite{siglip} &  Derm1M  & 0.5113 & 0.7162 & 0.2718 & 0.3574 & 0.2875 & 0.4288 & 0.7649 & 0.6867 & 0.7258 \\
          CoCa/ViT-B-32 \cite{coca} &  Derm1M  & 0.5635 & 0.6471 & 0.2019 & 0.2862 & 0.1942 & 0.3786 & 0.6431 & 0.6289 & 0.6360 \\
        \hline
        CLIP \cite{clip} &  Derm1M$^{\dagger}$ & 0.5631 & 0.7435 & 0.2559 & 0.3189 & 0.2485 & 0.4260 & 0.7348 & \textbf{0.6900} & 0.7124 \\
        SigLIP \cite{siglip} &  Derm1M$^{\dagger}$ & 0.5892 & 0.5282 & 0.2369 & 0.3045 & 0.2385 & 0.3795 & 0.6860 & 0.5701 & 0.6281 \\
       CoCa/ViT-B-32 \cite{coca} &  Derm1M$^{\dagger}$ & 0.5688 & 0.6911 & 0.2040 & 0.2557 & 0.2289 & 0.3897 & 0.6715 & 0.6408 & 0.6562 \\
        \rowcolor{gray!20}
        \textbf{MAKE (Ours)} & Derm1M$^{\dagger}$ & 0.5953 & \textbf{0.8266} & \textbf{0.3242} & \textbf{0.3914} & \textbf{0.3270} & \textbf{0.4929} & \textbf{0.7873} & 0.6864 & \textbf{0.7369} \\
        \bottomrule
    \end{tabular}}
    \label{tab1}

\end{table}

\begin{table}[t]
    \centering
     \small 
    \caption{Cross-modal retrieval performance comparison on the SkinCAP dataset. }
      \vspace{-2mm}
    \resizebox{0.75\textwidth}{!}{
    \begin{tabular}{l|c|ccc|ccc|c}
        \toprule
        \multirow{2}{*}{\textbf{Method}} & \multirow{2}{*}{\textbf{Pretrain Data}} & \multicolumn{3}{c|}{\textbf{Image-to-Text}} & \multicolumn{3}{c|}{\textbf{Text-to-Image}} & \multirow{2}{*}{\textbf{Average}} \\
        
        &  & R@10 & R@50 & R@100 & R@10 & R@50 & R@100 & \\
        \hline
        CLIP \cite{clip} & CLIP400M & 0.0913 & 0.2354 & 0.3407 & 0.0592 & 0.1860 & 0.2760 & 0.1981 \\
        BiomedCLIP \cite{biomedclip} & PMC-15M & 0.1359 & 0.3429 & 0.4698 & 0.1238 & 0.3304 & 0.4578 & 0.3101\\
        PMC-CLIP \cite{pmc-clip} & PMC-OA & 0.0672 & 0.1908 & 0.2783 & 0.0649 & 0.1855 & 0.2652 & 0.1753 \\
        MONET \cite{monet} & PubMed+TextBook & 0.1421 & 0.3384 & 0.4568 & 0.1492 & 0.3490 & 0.4756 & 0.3185 \\
        \hline
        CLIP \cite{clip} & Derm1M & 0.1532 & 0.3590 & 0.4851 & 0.1552 & 0.3715 & 0.4728 & 0.3328 \\
        SigLIP \cite{siglip} & Derm1M & 0.1757 & 0.3783 & 0.4871 & 0.1843 & 0.3896 & 0.4974 & 0.3521 \\
         CoCa/ViT-B-32 \cite{coca} & Derm1M & 0.1193 & 0.2865 & 0.3833 & 0.1291 & 0.3078 & 0.4089 & 0.2723 \\
        \hline
        CLIP \cite{clip} & Derm1M$^{\dagger}$ & 0.1587 & 0.3700 & 0.4788 & 0.1592 & 0.3537 & 0.4675 & 0.3313 \\
        SigLIP \cite{siglip} & Derm1M$^{\dagger}$ & 0.1835 & 0.3803 & 0.5006 & 0.1750 & 0.3793 & 0.4951 & 0.3523 \\
         CoCa/ViT-B-32 \cite{coca} & Derm1M$^{\dagger}$ & 0.1406 & 0.3241 & 0.4214 & 0.1429 & 0.3269 & 0.4202 & 0.2960 \\
        \rowcolor{gray!20}
        \textbf{MAKE (Ours)} & Derm1M$^{\dagger}$ & \textbf{0.2096} & \textbf{0.4440} & \textbf{0.5613} & \textbf{0.1995} & \textbf{0.4420} & \textbf{0.5628} & \textbf{0.4032} \\
        \bottomrule
    \end{tabular}}
    \label{tab2}
\end{table}
We compare our MAKE method with three groups of approaches across eight datasets on disease classification, concept annotation, and cross-modal retrieval. \textit{General VLMs} includes foundation models like BiomedCLIP \cite{biomedclip}, PMC-CLIP \cite{pmc-clip}, and MONET \cite{monet} trained on natural, biomedical, or dermatological image-text pairs. \textit{Standard VL methods} comprises SOTA vision-language approaches like CLIP \cite{clip}, SigLIP \cite{siglip}, and CoCa \cite{coca} pretrained on our Derm1M dataset. \textit{Knowledge-enhanced VL methods} includes the same approaches as Group 2 but pretrained on the knowledge-augmented version (Derm1M$^\dagger$). \\

\noindent\textbf{Zero-shot Skin Disease Classification and Concept Annotation:} Table~\ref{tab1} presents results across seven datasets for zero-shot disease classification and concept annotation. MAKE outperforms all methods on most datasets, with 7.58\% accuracy improvement on DermNet and 3.95\% on F17K over the best baseline. Overall, MAKE delivers 5.09\% higher average accuracy on classification and 1.11\% better AUROC on concept annotation than the best baseline. Three key findings emerge: (1) \textit{Standard VL Methods} trained on Derm1M substantially outperform \textit{General VLMs}; (2) VLMs trained on Derm1M$^\dagger$ often perform worse than on Derm1M, showing conventional VLMs cannot effectively utilize knowledge-augmented data; (3) Our MAKE framework achieves superior performance through multi-aspect knowledge contrastive learning framework\\

\noindent\textbf{Cross-modal Retrieval:} Table~\ref{tab2} evaluates VLMs' zero-shot image-text and text-image retrieval capabilities on the SkinCAP \cite{skincap} dataset. MAKE outperforms all baselines, achieving 44.4\% and 44.2\% R@50 for image-to-text and text-to-image retrieval, respectively. This represents a 6.57\% improvement over the best baseline (SigLIP on Derm1M$^{\dagger}$) for image-to-text and a 6.27\% gain for text-to-image retrieval, demonstrating MAKE's superior capability in aligning visual and textual representations in the dermatology domain.\\

\noindent\textbf{Ablation Study:} We perform ablation studies to analyze each component of our model, as shown in Table~\ref{tab3}. The first row represents our baseline, which uses the CLIP architecture. The second row shows CLIP with our multi-aspect knowledge-image contrastive learning loss (mkcl) using only \textit{knowledge set}, improving the baseline by 1.49\% in average accuracy. The third row incorporates mkcl with both \textit{knowledge set} and \textit{subtext set}, further improving performance by 2.07\% compared to mkcl with only \textit{knowledge set}, and by 3.56\% compared to the baseline. When adding local alignment loss (slra) to mkcl, the average accuracy further increases by 0.85\%. Finally, incorporating diagnosis knowledge-guided weighting (dkw) achieves the best performance of 49.29\%, which is 0.68\% higher than using only mkcl and slra, and 5.09\% higher than the baseline. These results demonstrate the effectiveness of all proposed components.

\begin{table}[t]
    \centering
\caption{Ablation study on different components of the MAKE framework. We report classification accuracy per dataset. $^\#$ denotes mkcl using only raw text, disease aspect, and concept aspect texts without spitted text for training.}
      \vspace{-2mm}
    \resizebox{0.6\textwidth}{!}{
    \begin{tabular}{ccc|ccccc|c}
        \toprule
        \multicolumn{3}{c|}{\textbf{Module}}& \multicolumn{6}{c}{\textbf{Disease Classification (ACC)}} \\
        \hline
        \textbf{mkcl} & \textbf{slra} & \textbf{dkw} & \textbf{PAD} & \textbf{DermNet} & \textbf{F17K} & \textbf{SD-128} & \textbf{SNU-134} & \textbf{Average} \\
        \midrule
          &  &  & 0.5957 & 0.7508 & 0.2595 & 0.3349 & 0.2689 & 0.4420 \\
        \checkmark$^\#$ &  &  & 0.6023 & 0.7439 & 0.3142 & 0.3465 & 0.2775 & 0.4569 \\
        \checkmark  &  &  & \textbf{0.6062} & 0.8093 & 0.3210 & 0.3533 & 0.2984 & 0.4776 \\
        \checkmark & \checkmark &   & 0.5653 & 0.8216 & 0.3191 & \textbf{0.4019} & 0.3227 & 0.4861 \\
        \rowcolor{gray!20} \checkmark & \checkmark & \checkmark  & 0.5953 & \textbf{0.8266} & \textbf{0.3242} & 0.3914 & \textbf{0.3270} & \textbf{0.4929} \\
        \bottomrule
    \end{tabular}}%
    \label{tab3}
\end{table}

\section{Conclusion}
In this paper, we introduce MAKE, a Multi-Aspect Knowledge-Enhanced vision-language pretraining framework for dermatology that addresses limitations of conventional medical VLP models through three innovations: multi-aspect knowle-dge-image contrastive learning, fine-grained alignment, and diagnosis-guided weig-hting. Experiments on diverse benchmarks demonstrate our framework's superior performance over SOTA VLP models. We hope our work inspires further research on multi-aspect knowledge-enhanced vision-language pretraining for medical domains.

\bibliographystyle{splncs04}
\bibliography{main}

\end{document}